# Deep Learning based Pedestrian Detection at Distance in Smart Cities

Ranjith K Dinakaran[1], Philip Easom[1], Ahmed Bouridane[1], Li Zhang[1], Richard Jiang[3], Fozia Mehboob[2] and Abdul Rauf[2]

[1] Computer and Information Sciences, Northumbria University, Newcastle upon Tyne, UK
[2] Computer Science, Imam Mohammed ibn Saud Islamic University, Kingdom Of Saudi Arabia
[3] The School of Computing and Communication, Lancaster, UK
Ranjith.Dinakaran@northumbria.ac.uk

**Abstract.** Generative adversarial networks (GANs) have been promising for many computer vision problems due to their powerful capabilities to enhance the data for training and test. In this paper, we leveraged GANs and proposed a new architecture with a cascaded Single Shot Detector (SSD) for pedestrian detection at distance, which is yet a challenge due to the varied sizes of pedestrians in videos at distance. To overcome the low-resolution issues in pedestrian detection at distance, DCGAN is employed to improve the resolution first to reconstruct more discriminative features for a SSD to detect objects in images or videos. A crucial advantage of our method is that it learns a multi-scale metric to distinguish multiple objects at different distances under one image, while DCGAN serves as an encoder-decoder platform to generate parts of an image that contain better discriminative information. To measure the effectiveness of our proposed method, experiments were carried out on the Canadian Institute for Advanced Research (CIFAR) dataset, and it was demonstrated that the proposed new architecture achieved a much better detection rate, particularly on vehicles and pedestrians at distance, making it highly suitable for smart cities applications that need to discover key objects or pedestrians at distance.

**Keywords:** Deep Neural Networks, Object Detection, Smart Homecare, Smart Cities

## 1 Introduction

Locating pedestrians on streets has been a primary task in many smart cities applications, such as driverless cars using car cameras, forensic surveillance in smart cities, and hospital surveillance on patients, and so on. Recently, deep convolutional neural networks (CNNs) have shown very promising results for automated object detection in videos and images, particularly in the detection of deformable objects such as pedestrians or faces [1-4, 13-17].

However, it is yet very challenging while the real-world applications demand very high detection rates under critical conditions, such as the detection of pedestrians at distance. A miss detection, for example in driverless vehicles, can result in an unrecoverable disaster in these applications. To address this challenge, in this paper, we report a new architecture that was designed for object detection at distance. While objects at distance are usually blurred in images or videos, deep convolutional generative adversarial networks (DCGANs) have shown a special power on improving the quality of images or videos from low resolution [5-12], due to the "generative" nature of DCGANs. By learning from realistic data, DCGANs can produce faked images or videos that are highly close to real data. Taking the advantages of DCGAN, we can easily employ it to cope with pedestrian detection at distance, which are usually challenging in the detection due to its small size and vague visual features.

Based on this assumption, we proposed a new architecture for pedestrian detection based on cascading a DCGAN with a Single Shot Detector (SSD) [4]. As a wider context for the work, using a combination of encoder-decoder based DCGANs cascaded with SSD can be utilized to extract the correct templates of objects at distance in an image or a video, where the GAN plays as a feature extractor to re-build the more realistic features of objects at distance to achieve a higher resolution for the following detection stage. Our contributions in this work may include:

1. Presenting a new architecture for object-detection at distance by combining DCGANs with SSDs.
2. Experimentally test if DCGAN-based enhancement on features can help improve the detection rates from SSDs.
3. Develop a practical method for the pedestrian detection at distance, which is highly demanded by real-world smart cities applications such as driverless cars.

Our evaluation is carried out via the quantified cross validation on the Canadian Institute for Advanced Research (CIFAR) dataset, and also illustrated by real-world on street videos for qualitative comparison.

## 2  Related Work

Object detection is considered as a major challenge in computer vision getting some success in recent years, thanks to recent advances of deep learning technologies. Among different deep learning architectures, GANs are one of most interesting architectures. As shown in Fig.1, the standard architectures of GANs often consist of two parts: one is a generator that can produce high-resolution images from low-resolution inputs, and the second is a discriminator that copes with the capability to examine the quality of produced high-resolution images and see if it is as realistic as possible.

Within various deep learning-based object detectors, You Only Look Once (YOLO) and SSD [4] are the two state-of-the-art methods that aim to capture the object regions in a very fast way. In this work, we have chosen SSD as our object detector, and cascaded it with a DCGAN generator (half of a DCGAN), as shown in Fig.2. The proposed combination aims to improve the detection rates of distant objects in images or videos by simply taking advantage of the DCGAN generator to produce higher resolution features of the objects.

The pipeline of the whole architecture in Fig.2 can be depicted as follows. The SSD, which is employed in this work, was first trained on ImageNet. As such, the SSD is basically a pretrained network in our experiment. The high-quality features reconstructed from DCGAN is then fed into the pretrained SSD, and the data propagates through the examiner's convolutional features from all layers, max-pooling each layer's representation to produce a 4×4 spatial grids. These features are then flattened and concatenated to form a 28672 dimensional vector and a regularized linear classifier is trained on top of them. Notably, the SSD detector performance lags in the scale factor, where the DCGAN can improvise the scale factor in SSD by providing the detector with super resolution images.

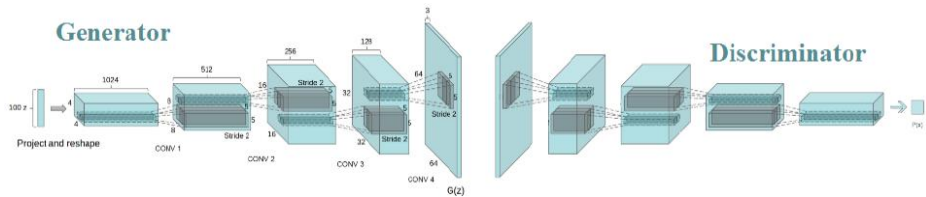

**Fig. 1.** Typical Architecture of DCGANs.

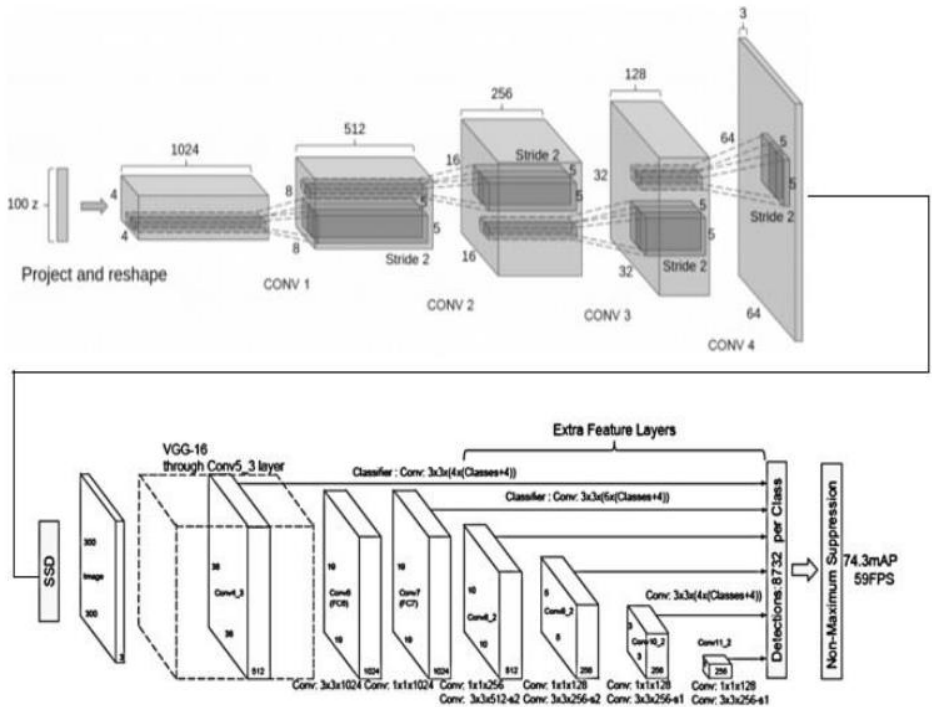

**Fig. 2.** The proposed DCGAN+SSD architecture for distant pedestrian detection.

In the convolutional detector, each feature layer is able to produce a fixed set of predictions for detection using convolutional filters as shown at top of the SSD architecture of Figure.2. The SSD is associated with a set of bounding boxes having default sizes relating to each cell from each feature map. The default size of bounding boxes is extracted in a convolutional manner with the feature map, so that the position of each bounding box is related to its sizes in the cell, and the class indicate the presence of feature cell in the bounding boxes.

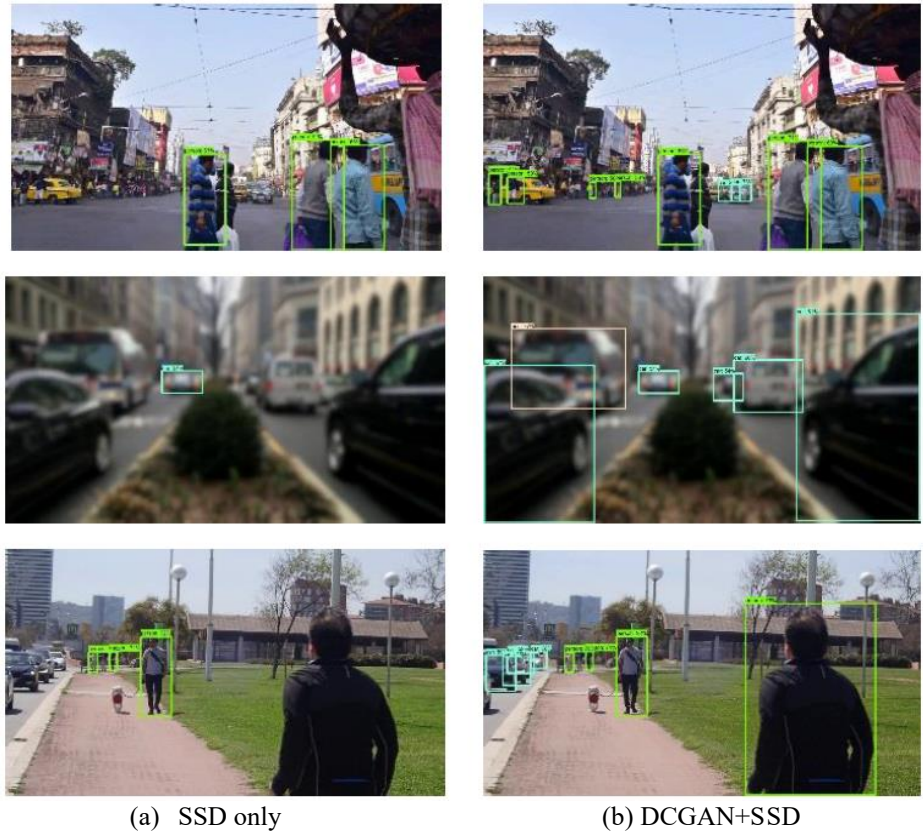

(a)  SSD only                                  (b) DCGAN+SSD

**Fig.3**. Detection examples for comparison. a) Results with SSD only; b) Results with DCGAN+SSD.

## 3    Experimental Results

In our experiments, we used CIFAR-10 dataset to carry out our validation and examine if the proposed GAN+SSD architecture outperforms the single SSD, particularly on distant object or pedestrian detection.

In our experiments, DCGAN was implemented in two steps. First, we implemented the DCGAN codes based on PyTorch and Tensorflow, as illustrated in [7-9], which recreates the state-of-the-art GAN results using multiple object encoding. We trained our DCGAN on the CIFAR-10 dataset, with all image size of 32×32, along with the batch size of 72, and 25 epochs, across a total of 60,000 images.

Fig.3 shows the detection results on the CIFAR-10 dataset, using DCGAN+SSD and SSD only, respectively. Fig.3-a) is the results from SSD only, where a number of objects were missed in the detection. Fig.3-b) shows the results from DCGAN+SSD, where a number of missed objects in Fig.3-a) were detected successfully. Particularly, objects at distance were detected in these images, which were mostly missed by the SSD only method.

Table 1 shows the detection rates on 100 images from CIFAR-10 and CIFAR-100 with distant objects in them. From the statistic results, it is clearly shown that our proposed DCGAN+SSD architecture can easily outperform the standard SSD, simply due to its improved detection rates on tiny objects at distance. The object detection rate was improved from 35.5% to 80.7%.

Table 1. Experimental Results for Comparison

| Method | Detection Rate |
| --- | --- |
| SSD Only | 35.5% |
| DCGAN+SSD | 80.7% |

## 4  Conclusion

In this work, we proposed a new architecture by cascading DCGANs with SSD to detect pedestrians and objects at distance, particularly for smart cities applications. With generative adversarial networks by implementing the DCGAN [6], more robust discriminative features are extracted around tiny objects and hence as a consequence, the detection rate is improved drastically. With such an apparent evidence to demonstrate its advantages, we can expect the proposed architecture will be valuable for smart cities applications that need to detect objects in the wild, particularly those tiny objects at distance, to secure life and avoid accidents. While GANs have been very successful in many other applications, our work successfully included it into a robust solution to the real world challenges in smart cities.